\documentclass[10pt,twocolumn,letterpaper]{article}

\usepackage[pagenumbers,applications]{wacv}

\usepackage{makecell}

\definecolor{wacvblue}{rgb}{0.21,0.49,0.74}
\usepackage[pagebackref,breaklinks,colorlinks,allcolors=wacvblue]{hyperref}

\def\wacvPaperID{*****}
\def\confName{WACV}
\def\confYear{2027}

\title{Frame-Consistent Feed-Forward 3D Gaussian Splatting from Sparse Unposed Images}

\author{Vinasirajan Viruthshaan\\
University of Moratuwa\\
{\tt\small viruthshaanv.20@uom.lk}
\and
Ranga Rodrigo\\
University of Moratuwa\\
{\tt\small ranga@uom.lk}
}

\begin{document}
\maketitle
Feed-forward Gaussian models have made novel view synthesis almost instantaneous. Feed-forward methods predict Gaussians directly, either in two stages or in a single merged pass. Using two separately trained models is attractive because each can specialize in its own objective: one for accurate pose estimation and the other for high-quality reconstruction, rather than forcing a single jointly trained network to compromise between the two. A natural instantiation pairs a frozen pose estimator, VGGT, with a frozen reconstruction backbone, TokenGS. Prior work, such as VGGT-X and 3R-GS, also adopts a two-stage design, but attributes the resulting quality degradation to noisy pose estimates and introduces learned refinement. We show that this is not the primary source of the problem. Through per-camera analysis across 107 scenes from two datasets, DL3DV and MipNeRF360, using COLMAP poses as the reference, we find that the rotation error after alignment is small (median $1.6^\circ$), whereas the translation scale is offset by a large, dataset-consistent factor. Controlled ablations over 140 DL3DV scenes show that the primary obstacles are two integration failures rather than pose error: a coordinate-frame inconsistency and a translation-scale mismatch, both of which can be corrected deterministically. With these fixes, a frozen VGGT driving a frozen TokenGS achieves $22.47$ PSNR, within $0.51$~dB of the COLMAP reference, with no additional training, when poses are estimated from all scene views. In the sparse setting, where reconstruction uses only four context images under the standard pose-free protocol, the same pipeline matches this all-view quality, yielding a fully COLMAP-free pipeline without any refinement step.
    
\section{Introduction}
\label{sec:introduction}

With 3D Gaussian Splatting (3DGS)~\cite{3dgs}, recent advances aim to make explicit reconstruction faster and more reliable without degrading quality.

Feed-forward Gaussian methods~\cite{noposplat,tokensplat,kang2025selfsplat,pf3plat,uni3r} use feed-forward networks, MLPs and transformers to predict Gaussians directly, with small or no refinement iterations, which makes reconstruction faster. Conventional 3DGS~\cite{3dgs}  instead does reconstruction in two stages: during the first stage, COLMAP~\cite{colmap} estimates the poses, during the second stage it initializes the Gaussians and refines them with iterative optimization.

Recent feed-forward methods fall into two families by how the pose and reconstruction components are trained. One family trains a \emph{single network end-to-end}: NoPoSplat~\cite{noposplat}, AnySplat~\cite{anysplat}, TokenSplat~\cite{tokensplat}, SelfSplat~\cite{kang2025selfsplat}, PF3plat~\cite{pf3plat}, and Uni3R~\cite{uni3r} jointly learn geometry and Gaussian prediction, so that pose estimation and reconstruction are optimized together within one model. The other family \emph{composes separately-trained models}: VGGT-X~\cite{vggt_x} and 3R-GS~\cite{3rgs} take poses from a pretrained geometry estimator and add a distinct learned refinement or optimization stage on top. Our work belongs to the latter family structurally, but connects two frozen pretrained models through deterministic operations alone, with no refinement or training.

Given images and poses, TokenGS~\cite{tokengs} unlike other feed-forward GS methods, predicts a fixed number of Gaussians, without other models that change the number of Gaussians based on image content. Its transformer architecture uses poses as Plücker ray embeddings to fuse with the image patches. But it requires already estimated poses, which is a blocker for using the model as a direct feed-forward pipeline that predicts Gaussians from unposed images. An easy integration is to use a direct pose estimator such as VGGT~\cite{VGGT} , MASt3R~\cite{MASt3R}, or DUSt3R~\cite{DUSt3R} to predict the pose. In this way we can use two separately trained large models, each for its specific objective: one for pose accuracy, one for good reconstruction. However, directly integrating the estimated context view poses degrades reconstruction quality substantially~\cite{vggt_x, 3rgs}, which raises concerns about the pose quality of these pose estimators.

Prior work largely addresses the degradation from imperfect geometric initialization through learned or optimization-based refinement. VGGT-X~\cite{vggt_x} keeps the VGGT~\cite{VGGT} predictions fixed and applies adaptive global alignment using XFeat correspondences, jointly refining the camera parameters with an epipolar loss, followed by MCMC-based 3DGS training with joint pose optimization. 3R-GS~\cite{3rgs} initializes from MASt3R-SfM~\cite{MASt3R-SfM} poses and addresses their imperfect initialization through MCMC-based pose optimization, an MLP-based global pose refiner, and a rendering-free epipolar geometric constraint. VGGS~\cite{vggs}, in contrast, leverages VGGT's multi-view geometric priors for sparse-view surface reconstruction, using an anchor-calibrated depth estimation scheme and a relative-depth consistency loss to align VGGT's depth predictions to the underlying surface. While it also contends with misalignment in VGGT's outputs, its focus is depth-level surface fidelity rather than the coordinate-frame and scale mismatch we identify when composing VGGT poses with a reconstruction backbone.

Instead, we investigate the actual source of this mismatch, rather than fixing it implicitly with a refinement network. Through controlled analysis, we find that the failure is not caused by pose inaccuracy. After alignment, VGGT's rotation error is small, while its translations are offset from the backbone's expected distribution by a large, systematic scale factor. Our controlled ablation over 140 DL3DV scenes shows that the major obstacles to merging VGGT and TokenGS are not pose errors, but two integration failures: coordinate-frame inconsistency and translation-scale mismatch, both of which admit deterministic fixes. With frame consistency and scale correction, we compose two frozen pretrained models (VGGT and TokenGS) with no retraining or fine-tuning. When VGGT estimates poses from all scene views, the composition reaches within 0.51 dB of the COLMAP-initialized baseline. The same fixes hold in the sparse setting, where VGGT estimates poses from only the context views, maintaining quality with no per-scene optimization.

In summary, our contributions are:
\begin{itemize}
    \item We \textbf{diagnose two integration failures in two-stage feed-forward 3DGS systems}: coordinate frame inconsistency and translation scale mismatch, and show via controlled ablation that these, not pose error, cause the quality collapse when connecting frozen feed-forward pose and 
    reconstruction models.

    \item We provide a \textbf{per-camera error decomposition} across 107 scenes (DL3DV~\cite{dl3dv}, MipNeRF360~\cite{mipnerf360}): rotation error is small (median $1.6^\circ$), while translation scale is offset by a large, dataset-consistent factor (approximately $7\times$).

    \item We introduce \textbf{deterministic, training-free fixes}, frame consistency and scale correction, reaching $22.47$ PSNR on 140 DL3DV~\cite{dl3dv} scenes, within $0.51$~dB of the COLMAP~\cite{colmap} reference, with poses estimated from all scene views, while maintaining the same inference speed.

    \item We show the fixes \textbf{hold in the sparse setting}: whereas the result above estimates poses from all scene views, here VGGT processes only the 4 context views together with one target at a time (the standard pose-free protocol), yet reconstruction quality is maintained, giving a fully COLMAP-free pipeline.
\end{itemize}
\section{Related Work}
\label{sec:relatedwork}
\subsection{Radiance Field Representation}

\begin{figure*}[t]
    \centering
    \includegraphics[width=\linewidth]{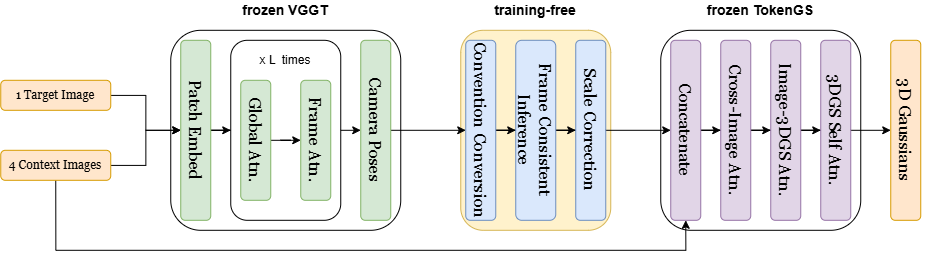}
    \caption{\textbf{Our training-free composition pipeline.} 
    A frozen VGGT estimates camera poses, and a frozen TokenGS 
    reconstructs 3D Gaussians; the two are connected by three 
    deterministic, training-free operations. All input views 
    (the four context images and the target) pass through VGGT 
    in a single forward pass so that every camera pose lies in 
    one shared coordinate frame. The predicted poses are then 
    corrected by convention conversion, frame-consistent 
    inference, and scale correction before conditioning TokenGS. 
    Reconstruction uses only the four context images (bottom 
    path); the target image is included solely to estimate its 
    pose in the shared frame. No component is trained or 
    fine-tuned.}
    \label{fig:pipeline}
\end{figure*}

Radiance field reconstruction has developed along two main lines: implicit representations such as NeRF~\cite{nerf}, and explicit ones such as 3DGS~\cite{3dgs}. Both typically rely on COLMAP~\cite{colmap} to initialize camera poses. Given the posed images, an implicit network is trained to represent the scene, or explicit primitives learn features that represent it, in both cases optimized independently for each scene.

Because this per scene optimization is slow, reconstructing a scene takes considerable time, which is prohibitive for real time applications. Subsequent work reduces this cost through faster optimization and compact representations~\cite{scaffoldgs, lightgaussian, minisplatting}, but the dependence on COLMAP~\cite{colmap} poses and per scene optimization remains. With the advancement of transformers, several methods predict pose and reconstruction in a single forward pass. We review these methods in the following subsections.
\subsection{Feed-Forward Pose and Geometry Estimation}

As networks become more powerful, 3D pose estimation can be solved directly by a neural network, almost entirely removing the time-consuming post-processing. DUSt3R~\cite{DUSt3R}, MASt3R~\cite{MASt3R}, and VGGT~\cite{VGGT} have shown results comparable to SfM based methods~\cite{colmap, sfm, vggsfm}.

However, DUSt3R~\cite{DUSt3R} and MASt3R~\cite{MASt3R} process only two images at a time and rely on post-processing to reconstruct larger sets, fusing pairwise reconstructions into a global one. VGGT~\cite{VGGT} instead processes one, a few, or even hundreds of views jointly in a single forward pass, achieving higher pose accuracy and often outperforming optimization-based alternatives without further processing.

We build on VGGT as a state-of-the-art feed-forward pose estimator; independent evaluations~\cite{wu2025evaluation} report its reliability in sparse-view settings. VGGT~\cite{VGGT} expresses all cameras in the coordinate frame of the first input view and is trained with scale-normalized supervision: two properties central to the integration failures we analyze in Section~\ref{sec:method}.

\subsection{Feed-Forward Gaussian Reconstruction}

A growing body of work builds Gaussian reconstruction on top of feed-forward geometry models~\cite{VGGT, MASt3R, DUSt3R}. These methods fall into two groups by how they use poses.

\textbf{Unified feed-forward networks.} One group trains a single network end-to-end to output Gaussians directly from unposed images. AnySplat~\cite{anysplat} adopts a VGGT-inspired geometry transformer and retrains it to jointly predict Gaussians and camera poses, while TokenSplat~\cite{tokensplat} trains a unified architecture with learnable camera tokens and an asymmetric decoder for joint pose and Gaussian estimation. SelfSplat~\cite{kang2025selfsplat} learns pose and Gaussian prediction jointly in a self-supervised manner, using a matching-aware pose network and depth refinement to couple geometry and reconstruction without ground-truth poses.

\textbf{Pose estimator with per-scene optimization or refinement.} A second group takes geometry or poses from a feed-forward estimator and then applies classical iterative optimization or learned refinement. VGGT-X~\cite{vggt_x} augments VGGT with adaptive global alignment (via XFeat correspondences and an epipolar objective) and robust MCMC-3DGS training for dense-view synthesis, and 3R-GS~\cite{3rgs} adds an MLP-based global pose refiner and a rendering-free epipolar constraint to MASt3R-SfM initializations. Both attribute the difficulty to imperfect or noisy pose estimates: VGGT-X frames VGGT's outputs as degrading initialization-sensitive training and reports that a gap to COLMAP persists even after adaptive alignment. VGGS~\cite{vggs} instead uses VGGT's geometric priors for a different task, sparse-view surface reconstruction, calibrating depth against multi-view anchors with a relative-depth consistency loss; its concern is depth-level surface fidelity rather than the coordinate-frame and scale mismatch we identify.

Common to these methods is the assumption that feed-forward geometry cannot be used as is: unified networks resolve this by retraining, while the second group relies on per scene optimization or learned refinement on top of a separate estimator. Whether two independently pretrained feed-forward models: a pose estimator and a reconstruction backbone, can be connected directly and kept frozen has not been studied. We show that they can. The obstacles are coordinate frame inconsistency and translation scale mismatch, and once these are addressed deterministically, frozen VGGT poses drive a frozen TokenGS to within 0.51~dB of the COLMAP reference with no training, and the same fixes hold in the sparse setting.

\section{Preliminaries}
\label{sec:preliminaries}

\subsection{Feed-Forward Pose Estimation with VGGT}

Given $N$ input images $\{I_i\}_{i=1}^{N}$, VGGT~\cite{VGGT} jointly regresses camera extrinsics $\{E_i\}_{i=1}^{N}$, camera intrinsics $\{K_i\}_{i=1}^{N}$, dense depth maps, and point maps. It tokenizes each image with a DINOv2~\cite{dinov2} encoder and appends a learnable camera token to each frame's tokens. These tokens are processed through alternating frame-wise and global attention layers, after which the attended camera tokens are decoded into camera poses by a separate head. In this work, we use only VGGT's predicted camera parameters.

Two properties of VGGT are central to our analysis. First, VGGT predicts all camera extrinsics, point maps, and depth maps in the coordinate frame of the first input camera: the first camera is assigned the identity pose (zero translation, no rotation), and all other cameras are expressed relative to it. This internal reference frame is unrelated to the world frame of any external reconstruction such as COLMAP~\cite{colmap}. Second, VGGT is trained with scale normalized supervision: each training scene's ground truth is normalized by the average distance of its 3D points to the origin, so that camera translations and geometry share a canonical unit scale. Rather than applying this normalization as a post-processing step, VGGT is trained to emit predictions already in this canonical scale. As a consequence, the scale of VGGT's predicted translations is tied to its internal normalization rather than to any external reference: the root cause, as we show in Section~\ref{sec:poseanalysis}, of the translation scale mismatch when VGGT poses are supplied to a reconstruction backbone trained on differently scaled poses.

\subsection{Feed-Forward Reconstruction with TokenGS}
\label{sec:tokengs}

Because its poses come from structure-from-motion, TokenGS implicitly assumes camera translations follow the coordinate convention and scale distribution of SfM reconstruction. This assumption is not unique to TokenGS: any feed-forward reconstruction backbone trained on posed images inherits the frame convention and scale of its training poses. When poses from a different source, with a different convention and scale---such as VGGT's internally normalized output---are supplied at inference, they may fall outside this expected distribution even when geometrically correct. We use TokenGS as a representative, state-of-the-art instance of this class; the integration failures we analyze arise from the general mismatch between how feed-forward pose estimators and reconstruction backbones define coordinate frames and scale.

\begin{figure*}[t]
    \centering
    \includegraphics[width=\linewidth]{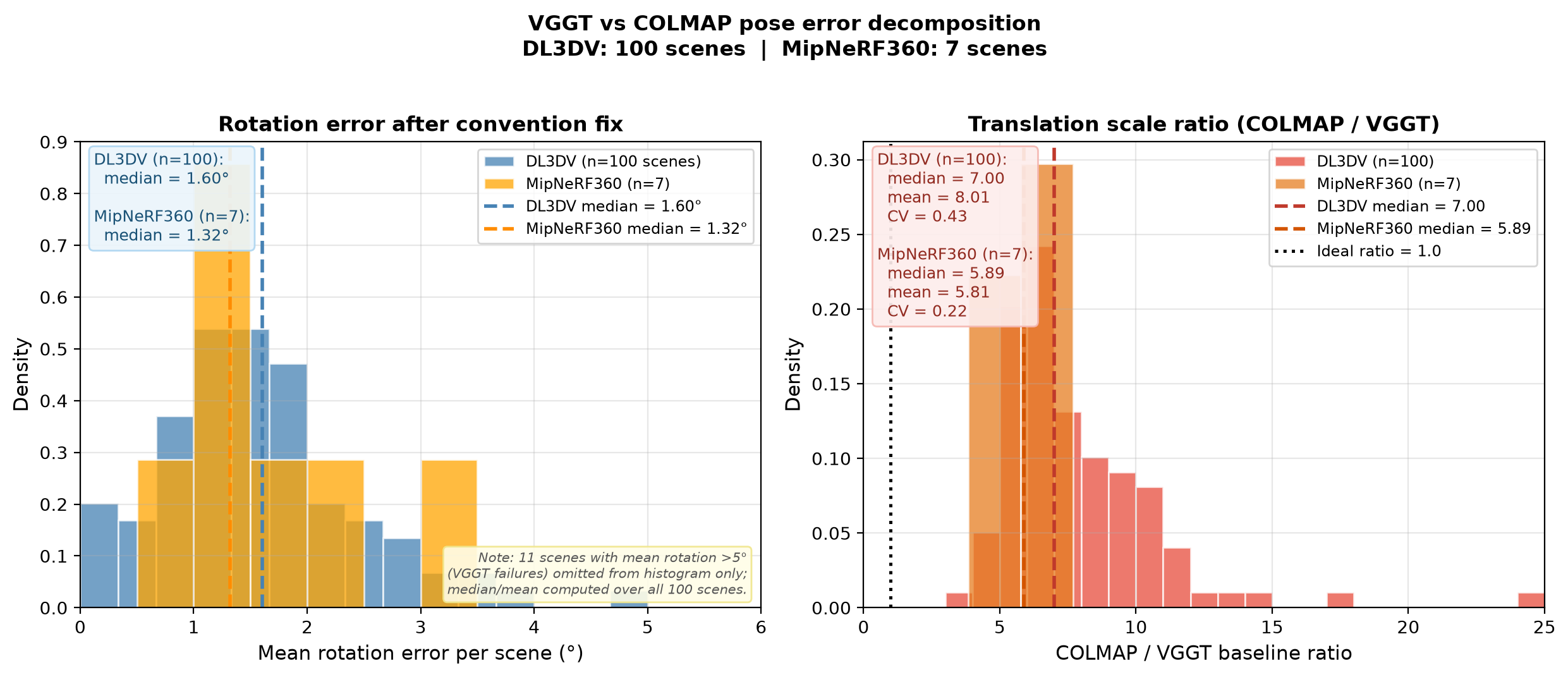}
    \caption{VGGT vs.\ COLMAP pose error decomposition. 
    Left: rotation error after convention alignment is small 
    (median $1.60^\circ$ on DL3DV, $1.32^\circ$ on MipNeRF360). 
    Right: the translation scale ratio is large and 
    consistent (median $7.0$ on DL3DV), far from the ideal 
    ratio of $1.0$. For typical scenes, rotation error is small; the dominant systematic offset is scale.}
    \label{fig:decomposition}
\end{figure*}

\section{Method}
\label{sec:method}

\subsection{Pose Analysis} 
\label{sec:poseanalysis}
\paragraph{Rotation Analysis via Procrustes Alignment.} 
VGGT and COLMAP predict camera poses in independent world frames, so their extrinsics cannot be compared directly. To isolate per-camera rotation error, we first align the two pose sets with a similarity transform (Sim(3): a shared scale, rotation, and translation) fitted by Procrustes alignment on the camera centers, then measure rotation after this alignment.

Given VGGT camera-to-world poses $\{P_i^{V}\}_{i=1}^{N}$ and COLMAP poses $\{P_i^{C}\}_{i=1}^{N}$, we solve for the similarity transform $(s,\mathbf{R},\mathbf{t})$ that best aligns the VGGT camera centers to COLMAP:
\begin{equation}
\label{eq:sim3}
\min_{s,\mathbf{R},\mathbf{t}}
\sum_{i=1}^{N}
\left\|
s\mathbf{R}\mathbf{c}_i^{V}
+\mathbf{t}
-\mathbf{c}_i^{C}
\right\|^{2},
\end{equation}

where scale $s \in \mathbb{R}^{+}$, rotation $\mathbf{R} \in SO(3)$, translation $\mathbf{t} \in \mathbb{R}^{3}$ and $\mathbf{c}_i=P_i[:3,3]$ is the $i$-th camera center. Eq.~(\ref{eq:sim3}) is solved in closed form by the Umeyama
algorithm~\cite{umeyama}, which returns the optimal
$(s,\mathbf{R},\mathbf{t})$ from the two sets of camera centers. After solving for $(s,\mathbf{R},\mathbf{t})$, we apply the recovered rotation $\mathbf{R}$ to each VGGT camera rotation ($\mathbf{R}_i^{V}$) to bring it into the COLMAP frame:
\begin{equation}
\tilde{\mathbf{R}}_i^{V} = \mathbf{R}\mathbf{R}_i^{V},
\end{equation}
and measure the residual per-camera rotation error as the geodesic distance
\begin{equation}
\theta_i =
\arccos\left(
\frac{
\operatorname{tr}\left(
\tilde{\mathbf{R}}_i^{V\top}\mathbf{R}_i^{C}
\right)-1
}{2}
\right).
\end{equation}
Here $\operatorname{tr}(\cdot)$ denotes the matrix trace, $\tilde{\mathbf{R}}_i^{V}$ denotes the aligned VGGT rotation and $\mathbf{R}_i^{C}$ is the corresponding COLMAP rotation. Crucially, this alignment is used \textit{only for measurement}; it is never part of our reconstruction pipeline, which requires no ground-truth poses.

Across 100 DL3DV and 7 MipNeRF360 scenes, the residual rotation error is small and consistent (median $1.60^\circ$ and $1.32^\circ$, respectively; Fig.~\ref{fig:decomposition}, left). TokenGS is explicitly designed to be robust to pose noise and multiview inconsistencies~\cite{tokengs}. Notably, the scale factor $s$ recovered by the same alignment is large (median $7.0$ on DL3DV), foreshadowing the translation scale mismatch we examine next.
\paragraph{Translation Scale Analysis.}

The Procrustes scale $s$ recovered in Eq.~\eqref{eq:sim3} already points to a
large VGGT--COLMAP translation mismatch. To quantify this discrepancy independently of the alignment, we compare camera baselines, the Euclidean distance between two camera centers, in the VGGT and COLMAP reconstructions. For each scene, we compute the ratio

\begin{equation}
r=
\frac{\left\|\mathbf{c}_1^{C}-\mathbf{c}_0^{C}\right\|}
{\left\|\mathbf{c}_1^{V}-\mathbf{c}_0^{V}\right\|},
\end{equation}
the COLMAP baseline divided by the VGGT baseline for the first two context cameras. A ratio of $r=1$ indicates matched scale, whereas $r>1$ indicates that VGGT translations are compressed relative to COLMAP.

Because individual baselines are sensitive to frame selection, we 
compute $r$ using well-separated context frames (quarter-point 
sampling) and report the per-scene distribution. Across the 100 
DL3DV training-split scenes, the ratio is consistent at the 
dataset level (median $7.0$), and across 7 MipNeRF360 scenes the 
median is $5.9$. While the per-scene spread is non-trivial (coefficient of variation $0.43$ on DL3DV, 
$0.22$ on MipNeRF360), a large offset appears in both datasets despite their different capture characteristics—well above the ideal ratio of $1.0$—indicating a systematic scale mismatch between the two pose conventions rather than a dataset-specific artifact.

This offset arises because VGGT and TokenGS use different scale conventions: VGGT's canonical scale differs from the SfM (COLMAP) scale that TokenGS expects, producing the measured factor of roughly seven. This large and consistent offset, compared with the relatively small rotation error, suggests that translation scale is the main source of failure when integrating the two models.

\paragraph{Robustness of the ratio estimate.}
A single camera pair can give a noisy baseline, so we examined the sensitivity of the ratio $r$ to frame selection across five strategies on 100 DL3DV scenes. Using \emph{adjacent} context frames yields highly unstable estimates (coefficient of variation $\mathrm{CV}=1.83$), because closely spaced cameras have small baselines that amplify noise. 

Using well-separated frames sharply stabilizes the estimate. Quarter-point
sampling gives a median ratio of $7.0$ ($\mathrm{CV}=0.43$), and taking the
median over all context camera pairs is the most robust ($\mathrm{CV}=0.39$).
Across all stable variants, the median ratio stays within $6.7$--$7.1$,
confirming that the scale offset is a stable scene-level property rather than
an artifact of frame choice.

\begin{table*}[t]
\centering
\caption{Main results on 140 DL3DV scenes (960$\times$540, DL3DV \texttt{images\_4}). Deterministic fixes bring frozen VGGT poses to within 0.51\,dB of the COLMAP ceiling with no training. All methods reconstruct Gaussians from the same four context views and the same image resolution; rows differ only in how poses are estimated. COLMAP uses the full capture ($\sim$350 frames). ``Joint eval views'' runs VGGT once on the 50 TokenGS eval frames (4 context $\cup$ 50 targets). ``Sparse, per-target'' runs VGGT on five views (4 context + one target) per novel view. End-to-end time is pose estimation + TokenGS.}
\label{tab:main}
\begin{tabular}{@{}lccccccc@{}}
\toprule
Pose method &
\#Views &
PSNR$\uparrow$ &
LPIPS$\downarrow$ &
SSIM$\uparrow$ &
\shortstack{Pose Est.\\Time $\downarrow$} &
\shortstack{TokenGS\\Time $\downarrow$} &
\shortstack{End-to-end\\Time $\downarrow$} \\
\midrule
COLMAP (ceiling) &
$\sim$350 &
22.98 &
0.328 &
0.737 &
$\sim$9\,min$^{\dagger}$ &
0.155\,s &
$\sim$9\,min \\
\midrule
VGGT (joint eval views) &
50 &
22.47 &
0.333 &
0.715 &
6.64\,s &
0.155\,s &
6.79\,s \\
VGGT (sparse, per-target) &
5 &
22.44 &
0.330 &
0.714 &
0.598\,s &
0.155\,s &
0.753\,s \\
\bottomrule
\end{tabular}
\begin{flushleft}
\footnotesize $^{\dagger}$Average COLMAP SfM time over 20 DL3DV scenes (all frames from
\texttt{images\_4}, $\sim$350 images/scene; $\sim$6\,min/scene).
Not timed on the full 140-scene eval.
\end{flushleft}
\end{table*}

\subsection{Deterministic Integration}
Our pipeline connects a frozen VGGT pose estimator to a frozen TokenGS reconstruction backbone through deterministic operations that require no learned components and no ground-truth poses. Section~\ref{sec:poseanalysis} identified two failure modes, which we address in turn. We resolve the coordinate frame inconsistency with two operations, a convention conversion (\S\ref{sec:convention}) and frame-consistent inference (\S\ref{sec:allframe}), and the scale mismatch with a single scale correction (\S\ref{sec:scale}). Figure~\ref{fig:pipeline} gives an overview.

\subsection{Coordinate Convention Conversion}
\label{sec:convention}

VGGT outputs camera extrinsics in a world-to-camera (W2C) convention following the OpenCV camera model ($x$-right, $y$-down, $z$-forward)~\cite{VGGT}. TokenGS, instead expects camera-to-world (C2W) poses in the NeRF/OpenGL convention ($x$-right, $y$-up, $z$-backward). We bridge these with two deterministic steps.

First, we invert each VGGT extrinsic $E=[\mathbf{R}\mid\mathbf{t}]$ to obtain a C2W pose $P=E^{-1}$, whose rotation is $\mathbf{R}^{\top}$ and camera center is $\mathbf{c}=-\mathbf{R}^{\top}\mathbf{t}$. Second, we convert the axis convention by negating the $y$ and $z$ axes of the rotation, i.e., right-multiplying the C2W rotation by $\mathbf{M}=\mathrm{diag}(1,-1,-1)$, yielding the final pose $P'$. This axis flip acts only on the rotation, leaving the camera center unchanged, so the convention conversion is orthogonal to the scale correction of Section~\ref{sec:scale}. 

\subsection{Frame Consistency}
\label{sec:allframe}
The dominant failure arises when context and target poses come from different coordinate frames (Section~\ref{sec:method}). Registering external target poses into the context frame via Sim(3) alignment is ill-conditioned on near-collinear captures. We instead place each target in the context frame by construction: the four context views and that target are passed through VGGT in a single forward pass, so all poses share one frame. Only the four context views are used for reconstruction; the target is included solely to obtain its pose. This follows the standard pose-free 
protocol and keeps reconstruction sparse while guaranteeing frame 
consistency, with no external alignment.

\textbf{Validation}. To measure the composition against COLMAP under matched information, we additionally run VGGT on \textbf{all scene views } in one pass. Context and target poses are then mutually consistent, and reconstruction from the four context views reaches within 0.51 dB of the COLMAP reference (Section~\ref{sec:experiments})—confirming that VGGT poses drive TokenGS correctly once frame and scale are fixed.

\textbf{Sim(3) offset cancels.}
Let $\mathbf{T} \in \operatorname{Sim}(3)$ be the unknown similarity transform between VGGT's world frame and COLMAP's. When the context poses (which build the Gaussians) and the target pose share a single VGGT frame, both are expressed under the same $\mathbf{T}$. Rendering depends only on the relative pose between the target camera and the Gaussian representation; since $\mathbf{T}$ applies identically to both, it cancels exactly in the relative transform. The reconstruction is therefore invariant to the absolute frame, and no alignment to COLMAP is required.

\subsection{Scale Correction}
\label{sec:scale}
After the convention and frame fixes, VGGT translations remain 
systematically smaller than TokenGS expects. VGGT normalizes each 
scene to its own canonical scale, based on the mean point-to-origin 
distance of its 3D points, whereas TokenGS was trained on 
SfM~(COLMAP)-scale poses; the two differ by a systematic 
factor.\footnote{TokenGS's DL3DV checkpoint expects input poses at 
SfM~(COLMAP) scale, which it rescales by a fixed $0.15$ for unit 
mean depth~\cite{tokengs,depthsplat}. Our factor brings VGGT's 
canonical scale up to this expected SfM input scale; TokenGS's own 
rescaling is left unchanged.} We calibrate 
a single correction factor of $7.0$ on 100 DL3DV training scenes, 
disjoint from the evaluation set, and apply it directly to every 
VGGT translation. On the 140 evaluation scenes, the per-scene ratio 
has median $6.86$ and mean $7.22$, confirming the constant is 
well-centered. The factor is dataset-specific (approximately $6.0$ 
on MipNeRF360) but requires no per-scene computation. It could 
equivalently be absorbed into TokenGS's fixed scale constant; we 
apply it externally to leave the backbone unmodified.
\section{Experiments}
\label{sec:experiments}

\begin{figure*}[t]
    \centering
    \includegraphics[width=\linewidth]{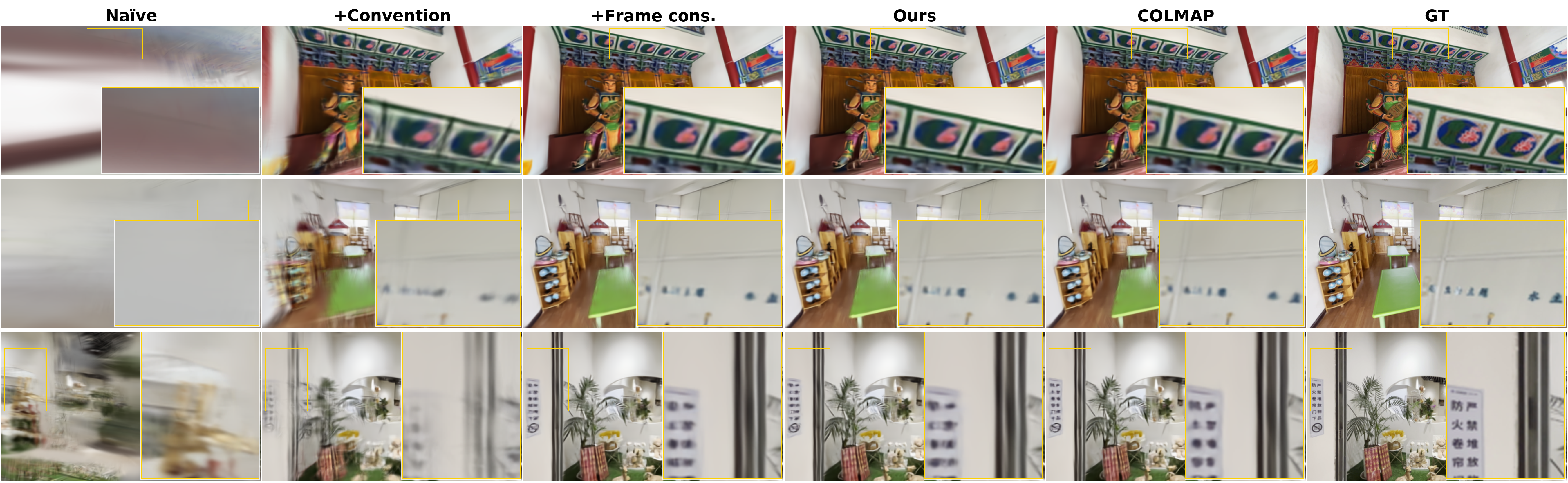}
    \caption{\textbf{Qualitative comparison on DL3DV.} From left to 
right: na\"ive integration, with convention conversion added, with 
frame-consistent inference added, our full method (with scale 
correction), the COLMAP-pose reference, and ground truth. 
\textbf{Insets show a magnified region for closer comparison.} 
Na\"ive integration collapses under the coordinate-frame mismatch; 
each deterministic fix progressively restores structure, and our 
full pipeline produces renderings visually indistinguishable from 
the COLMAP reference, using no ground-truth poses.}
\label{fig:qualitative}
\end{figure*}

\subsection{Setup}
\paragraph{Datasets.} We evaluate on the DL3DV~\cite{dl3dv} 
benchmark using the standard 4-view protocol, with context frames 
selected by the DepthSplat evaluation indices~\cite{depthsplat}. 
Our main results use 140 benchmark scenes; each scene provides 4 
context views and up to 50 target views for rendering. We evaluate 
end-to-end reconstruction on DL3DV because the TokenGS checkpoint 
is publicly released only for this dataset. For the pose and scale 
analysis (\S\ref{sec:poseanalysis}), we use a separate set of 100 DL3DV 
scenes drawn from the training split, disjoint from the 140 
evaluation scenes, and additionally 7 scenes from 
MipNeRF360~\cite{mipnerf360}. Our scale constant is calibrated on 
these training scenes and never on the evaluation set. The scale 
offset persists on MipNeRF360, indicating the failure modes are 
not DL3DV-specific.

\paragraph{Backbone and poses.} We use the publicly released TokenGS 
4-view checkpoint~\cite{tokengs} as a frozen reconstruction backbone 
and VGGT-1B~\cite{VGGT} as a frozen pose estimator. Neither model is 
finetuned. Camera poses are obtained either from COLMAP (upper-bound 
reference) or from VGGT with our integration fixes.

\paragraph{Pose settings.} We report two settings, both reconstructing 
the Gaussians from only the four context views. In the \emph{all scene 
views} setting, VGGT estimates poses from all scene views in a single 
forward pass; this validates the composition against COLMAP under 
matched information. In the \emph{sparse} setting, following the 
standard pose-free protocol, each target's pose is obtained from a 
forward pass over the four context views together with that target; 
the target is used only for its pose, and reconstruction remains sparse.

\paragraph{Metrics.} We report PSNR, SSIM~\cite{ssim}, and 
LPIPS~\cite{lpips} on rendered target views, averaged across all scenes, 
and end-to-end inference time on a single RTX 3090 GPU.

\paragraph{Baseline.} Our reference is the \emph{COLMAP ceiling}: since 
DL3DV provides COLMAP poses by default, this represents the quality 
achievable with an accurate but slow SfM pipeline. Our method replaces 
these poses with VGGT predictions plus deterministic fixes, targeting 
COLMAP-level quality without COLMAP and without additional training. 
We note that DL3DV's COLMAP poses are computed on the full dense image 
sequence, whereas our sparse setting uses only four context views.

\subsection{Main Results}
Table~\ref{tab:main} reports our main comparison on 140 DL3DV scenes. 
Supplying frozen VGGT poses to frozen TokenGS through our deterministic 
fixes reaches $22.47$~PSNR when poses are estimated from all scene 
views, within $0.51$~dB of the COLMAP ceiling ($22.98$). Our reproduced 
COLMAP ceiling matches TokenGS's own reported DL3DV performance 
($23.02$ without token tuning~\cite{tokengs}), confirming our evaluation 
faithfully reproduces the backbone. In the sparse setting, reconstructing 
from only four context views, the same fixes reach $22.44$~PSNR, 
essentially matching the all-view result. Both settings match the 
backbone's inference time of $\sim$$155$~ms per scene, since the fixes 
add only cheap deterministic operations on camera poses. Our method 
requires no ground-truth poses, no per-scene optimization, and no 
network fine-tuning.

\paragraph{Inference cost.} A single reconstruction runs one VGGT 
forward pass ($\sim$$0.6$~s) followed by TokenGS ($\sim$$155$~ms), 
sub-second end-to-end, versus COLMAP's minutes-to-hours for 
pose estimation on the full sequence. In the sparse setting, scoring 
all target views of a scene requires one VGGT pass per target for 
evaluation; this is an evaluation cost, not a deployment cost, as a 
single deployment reconstruction remains sub-second.

\subsection{Component Ablation}

\begin{table}[t]
\centering
\caption{Component ablation on 140 DL3DV scenes. The frame-consistency 
fix accounts for the majority of the recovery.}
\label{tab:ablation}
\begin{tabular}{@{}lc@{}}
\toprule
Configuration & PSNR$\uparrow$ \\
\midrule
Na\"ive integration (frame mismatch) & 13.67 \\
+ Convention conversion & 20.38 \\
+ frame-consistent inference & 22.24 \\
+ Scale correction ($\times 7.0$) & 22.47 \\
\midrule
COLMAP ceiling & 22.98 \\
\bottomrule
\end{tabular}
\end{table}

Table~\ref{tab:ablation} isolates the contribution of each fix. 
Starting from na\"ive integration, where context and target poses lie 
in different coordinate frames, quality collapses to $13.67$~PSNR. 
Resolving the coordinate-frame inconsistency through convention 
conversion and frame-consistent inference recovers the majority of 
quality, lifting PSNR by $8.57$ dB. Adding scale correction closes 
most of the remaining gap to COLMAP. This confirms that frame 
consistency---not pose refinement---is the dominant factor, and that 
both fixes are necessary.

\subsection{Intrinsics}
\begin{table}[t]
\centering
\small
\caption{VGGT intrinsics on 140 DL3DV scenes. Replacing COLMAP 
intrinsics with VGGT's, keeping COLMAP extrinsics, leaves quality 
essentially unchanged.}
\label{tab:intrinsic}
\begin{tabular}{@{}lc@{}}
\toprule
Configuration & PSNR$\uparrow$ \\
\midrule
COLMAP intrinsics + COLMAP extrinsics & 22.98 \\
VGGT intrinsics + COLMAP extrinsics & 22.94 \\
\bottomrule
\end{tabular}
\end{table}
VGGT predicts camera intrinsics alongside extrinsics. To confirm 
that its intrinsics are directly usable, we replace COLMAP's 
intrinsics with VGGT's while keeping COLMAP extrinsics: quality is 
essentially unchanged ($22.94$ vs.\ $22.98$). VGGT's intrinsics can 
therefore be used as-is. The remaining integration challenge lies 
in the extrinsics---not because VGGT's rotations or translations are 
inaccurate (our analysis shows they are; Section~\ref{sec:poseanalysis}), 
but because they must be expressed in a coordinate frame and scale 
consistent with the reconstruction backbone.

\subsection{Qualitative Results}
Figure~\ref{fig:qualitative} shows rendered novel views comparing 
na\"ive integration, our method, and the COLMAP ceiling. Na\"ive 
integration produces severe misalignment and ghosting caused by the 
frame inconsistency, while our fixes yield renderings visually 
indistinguishable from the COLMAP-initialized results, despite using 
no ground-truth poses.
\section{Conclusion}
We revisited the problem of connecting feed-forward pose estimation to feed-forward 3D reconstruction and showed that the apparent difficulty is not due to pose inaccuracy but to two overlooked, deterministically fixable integration failures: coordinate frame inconsistency and translation scale mismatch. With a convention conversion, all-frame inference, and a simple scale correction—no training, no ground-truth poses, no per-scene optimization—frozen VGGT poses drive a frozen TokenGS backbone to within 0.51~dB of the COLMAP-initialized ceiling at identical inference speed. Our per-camera and cross-dataset analyses establish that rotation error is small and scale offset is systematic, reframing a problem previously attributed to pose quality as one of integration design.

\paragraph{Why not joint training?}
A natural alternative is to train VGGT and TokenGS jointly, as in 
AnySplat~\cite{anysplat} and TokenSplat~\cite{tokensplat}. We 
deliberately avoid this: joint training is costly and, most 
importantly, would obscure the diagnosis. By keeping both models 
frozen, we isolate the integration failures and show they are 
fixable without any learning---a finding that transfers to future 
pose estimators and reconstruction backbones. Notably, even VGGT-X~\cite{vggt_x}, which applies learned 
adaptive alignment to VGGT outputs in the dense-view regime, 
reports that a gap to COLMAP persists after refinement. This is 
consistent with our diagnosis: if the difficulty were purely 
unrefined pose error, such refinement should close it. Our results 
instead point to coordinate-frame and scale mismatches as the 
source, which admit a deterministic correction.

\paragraph{Limitations.}
First, establishing a shared coordinate frame requires VGGT to 
process context and target views together. In the standard 
benchmark protocol, target viewpoints come from the same capture 
and are available to VGGT. For deployment with arbitrary 
user-specified viewpoints, the target must instead be placed in 
the context frame; because VGGT estimates poses jointly, the 
cleanest option is to include the target viewpoint in the same 
forward pass, rather than post-hoc registration, which our analysis 
shows is ill-conditioned on near-collinear captures. This affects 
only the \emph{specification} of the target viewpoint, not the 
reconstruction, which is built entirely from the context views. A 
full interactive implementation is left to future work.

Second, the fixed multiplier is dataset-specific ($\times 7.0$ on 
DL3DV, $\times 6.0$ on MipNeRF360); a per-scene normalization from 
context-view geometry would remove this dependence, which we leave 
to future validation.

Finally, our end-to-end evaluation uses a single reconstruction 
backbone (TokenGS) on a single dataset (DL3DV), constrained by 
checkpoint availability. Our pose-level analysis on MipNeRF360, 
where the same scale offset appears, suggests the failure modes are 
not backbone- or dataset-specific, but validating the fixes across 
other backbones end-to-end is left to future work.

We hope this diagnosis guides the practical, training-free combination of the increasingly capable feed-forward models now emerging.
{
    \small
    \bibliographystyle{ieeenat_fullname}
    \bibliography{main}
}

\end{document}